\documentclass[10pt,twocolumn,letterpaper]{article}

\usepackage{cvpr}              

\usepackage{graphicx}
\usepackage{amsmath}
\usepackage{amssymb}
\usepackage{booktabs}
\usepackage{subfiles}
\usepackage{multirow}
\usepackage[accsupp]{axessibility}
\usepackage[pagebackref,breaklinks,colorlinks]{hyperref}

\usepackage[capitalize]{cleveref}
\crefname{section}{Sec.}{Secs.}
\Crefname{section}{Section}{Sections}
\Crefname{table}{Table}{Tables}
\crefname{table}{Tab.}{Tabs.}

\def\cvprPaperID{2989} 
\def\confName{CVPR}
\def\confYear{2022}

\begin{document}

\title{MonoGround: Detecting Monocular 3D Objects from the Ground}


\author{Zequn Qin$^1$, Xi Li$^{1,2,3}$\thanks{Corresponding author.}
\\
$^1$College of Computer Science, Zhejiang University
\\
$^2$Shanghai Institute for Advanced Study of Zhejiang University; $^3$Shanghai AI Lab
\\
{\tt \small zequnqin@gmail.com, xilizju@zju.edu.cn}
}
\maketitle

\begin{abstract}
Monocular 3D object detection has attracted great attention for its advantages in simplicity and cost. Due to the ill-posed 2D to 3D mapping essence from the monocular imaging process, monocular 3D object detection suffers from inaccurate depth estimation and thus has poor 3D detection results. To alleviate this problem, we propose to introduce the ground plane as a prior in the monocular 3d object detection. The ground plane prior serves as an additional geometric condition to the ill-posed mapping and an extra source in depth estimation. In this way, we can get a more accurate depth estimation from the ground. Meanwhile, to take full advantage of the ground plane prior, we propose a depth-align training strategy and a precise two-stage depth inference method tailored for the ground plane prior. It is worth noting that the introduced ground plane prior requires no extra data sources like LiDAR, stereo images, and depth information. Extensive experiments on the KITTI benchmark show that our method could achieve state-of-the-art results compared with other methods while maintaining a very fast speed. Our code and models are available at \url{https://github.com/cfzd/MonoGround}.
\end{abstract}
\vspace{-15pt}
\section{Introduction}
\label{sec_intro}

3D object detection is a fundamental computer vision task that aims to obtain the locations, sizes, and orientations of objects. To get the real-world 3D information, many methods adopt modalities like point clouds from LiDAR, stereo images, and depth images, which require extra sensors of data sources. Different from them, monocular 3D object detection, which only requires a single 2D image and camera calibration information, increasingly draws the community's attention for its superiorities in simplicity and cost, especially in the autonomous driving field.

\begin{figure}
    \centering
    \begin{subfigure}{1.0\linewidth}
    \includegraphics[width=1.0\linewidth]{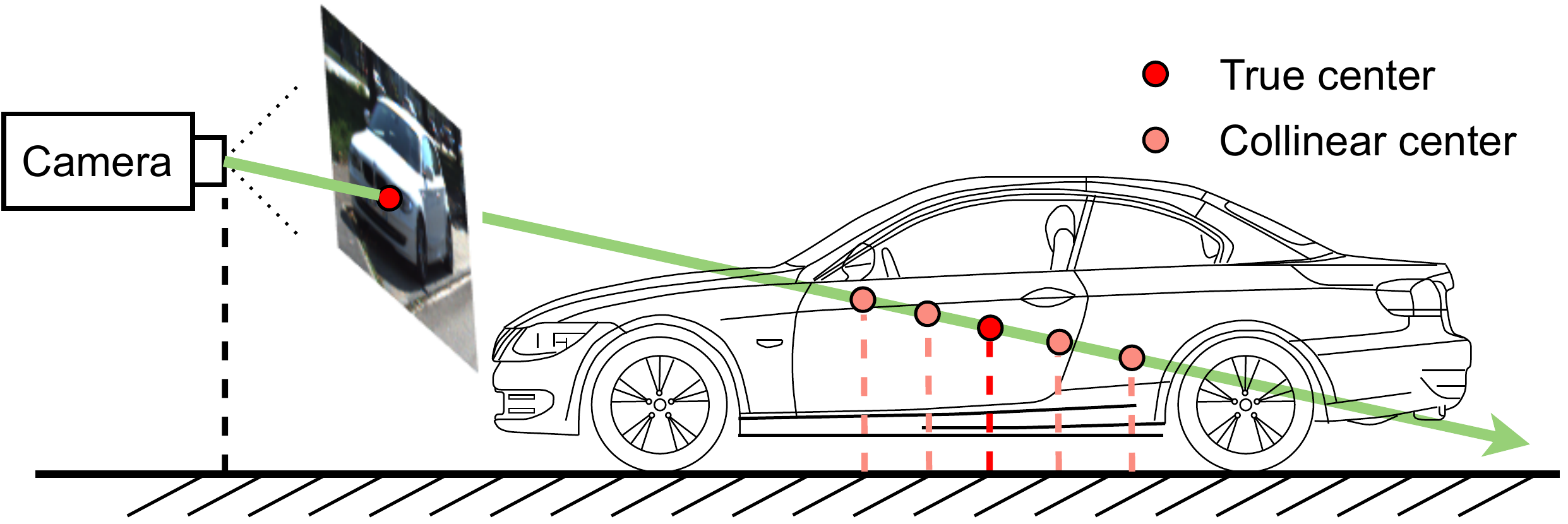}
    \caption{Multiple depth solutions with collinear points.}
    \label{fig_intro_collinear}
    \end{subfigure}
    \begin{subfigure}{1.0\linewidth}
    \includegraphics[width=1.0\linewidth]{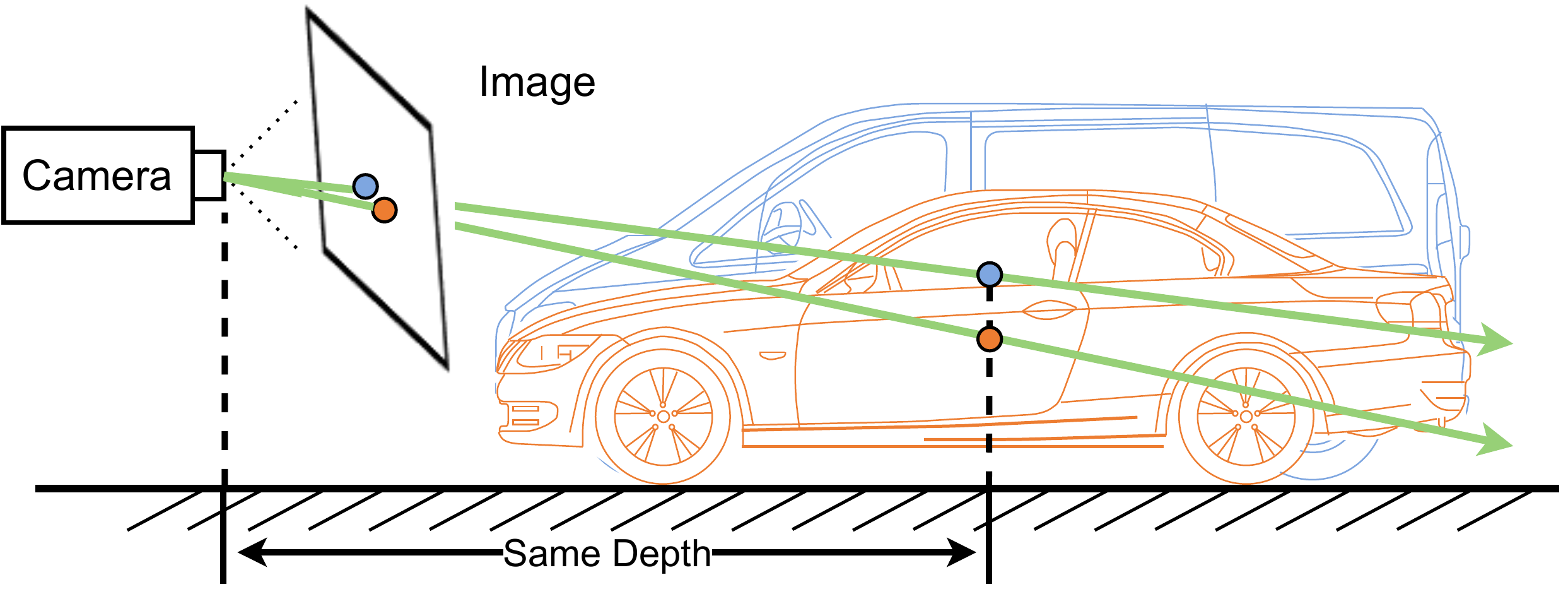}
    \caption{Two cars with the same depths and positions, but the projected centers are different in the image space. Depth is correlated with heights.}
    \label{fig_intro_differ}
    \end{subfigure}
    \vspace{-5pt}
    \caption{Difficulties in the monocluar 3D object detection.}
    \vspace{-15pt}
\end{figure}

Despite the superiorities of monocular 3D object detection, obtaining 3D information from a single 2D image is essentially hard for the following reasons: 1) the mapping from 2D to 3D is an ill-posed problem since a location in the 2D image plane corresponds to all collinear 3D positions, which are in the ray from the optical center to the 2D location. This one-to-many property makes predicting the depths of objects difficult, as shown in \cref{fig_intro_collinear}. 2) The expression of 3D object's depth is correlated with the height. For two objects with the same depths and positions, if the heights are different, the projected center would differ, as shown in \cref{fig_intro_differ}. In this way, the learning of depths has to overcome the interference with irrelevant attributes of objects. 3) Mainstream monocular 3D object detection methods~\cite{liu2020smoke,MonoFlex,lu2021geometry,zhou2021monocular} are commonly based on the CenterNet~\cite{zhou2019objects}. Under this framework, all information of each object is represented with a point, including the depth information. In this way, the supervision of depth is very sparse during training (Suppose there are two cars in an image, then only two points on the depth map would be trained, and all other points on the depth map are ignored). Such sparse supervision would lead to insufficient learning of depth estimation, which significantly differs from common monocular depth estimation tasks with dense depth supervision.

\begin{figure}
    \centering
    \includegraphics[width=1.0\linewidth]{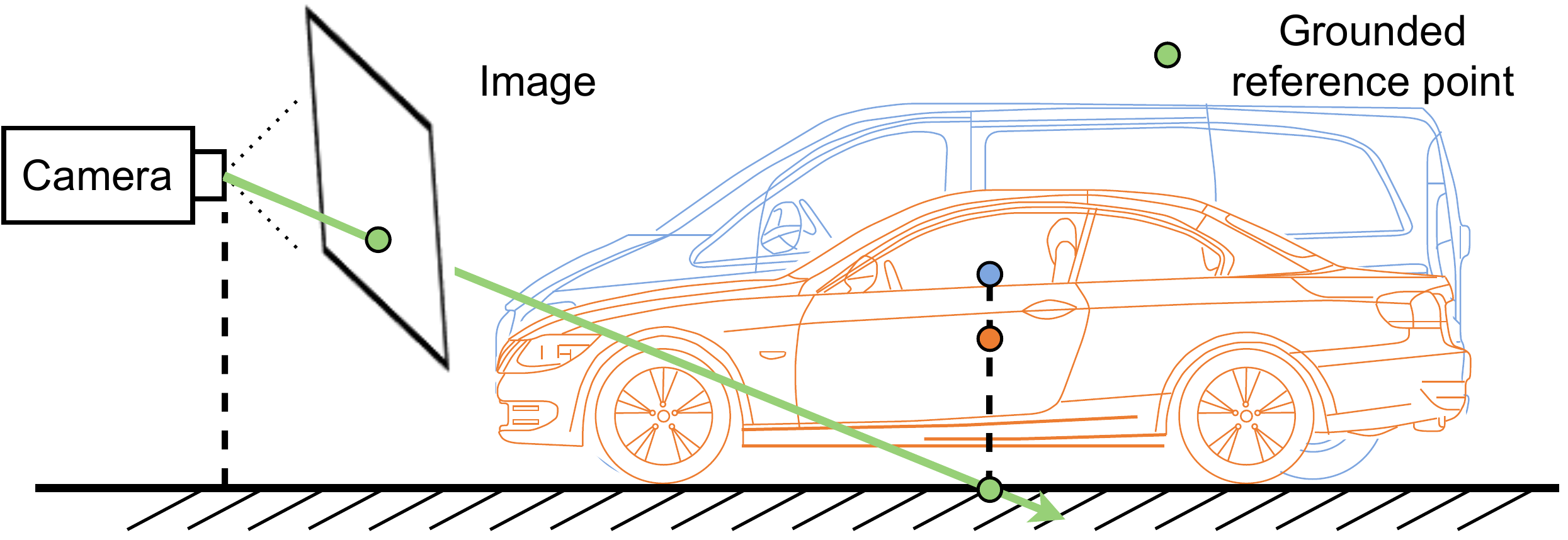}
    \caption{Illustration of the depth on the ground.}
    \label{fig_intro_ground}
\end{figure}

To address the above problems, we propose to introduce the ground plane prior to the monocular 3D object detection task. With the ground plane prior, the ill-posed 2D to 3D mapping becomes a well-posed problem with a unique solution. The unique solution is determined by the intersection of the camera ray and the ground plane. The expression of depth is no longer correlated with the object's height, as shown in \cref{fig_intro_ground}. Moreover, since the ground plane is introduced and utilized, we could expand the original sparse depth supervision to a dense depth supervision by using the dense depth from the ground plane.

With the above motivations, we propose our monocular 3D object detection model with the dense ground plane prior, termed as MonoGround. Within this formulation, we propose a depth-align method to effectively learn and predict dense grounded depth. Meanwhile, with the help of dense predicted depth, we also propose a two-stage depth inference method, which brings finer-grained depth estimation. In summary, the main contribution of this work can be summarized as follows:

\begin{itemize}
    \vspace{-2pt}
    \item We propose to introduce the ground plane prior to the monocular 3D object detection, which could alleviate ill-posed mapping, remove irrelevant correlation of depths, and provide dense depth supervision, without any extra data like LiDAR, stereo, and depth images.
    \vspace{-2pt}
    \item We propose a depth-align training strategy and a fine-grained two-stage depth inference method to take advantage of the introduced ground plane prior and achieve precise depth inference.
    \vspace{-2pt}
    \item Experiments on the KITTI dataset show the effectiveness of introducing the ground plane prior and the proposed method, and our method achieves the SOTA performance with a very fast speed in real-time.
\end{itemize}

\vspace{-10pt}
\section{Related Work}
\label{sec_related_work}
\paragraph{Monocular 3D Object Detection}
Monocular 3D object detection methods can be roughly divided into two groups. The first kind of method try to utilize extra available data sources to simplify the detection of 3D objects. The extra data sources can be LiDAR, depth images, and CAD models. The most commonly used extra data sources are the LiDAR~\cite{ku2018joint,shi2020pv,shi2019pointrcnn} and depth information~\cite{cai2020monocular,ma2019accurate,ma2020rethinking,ye2020monocular,ding2020learning,wang2021progressive,zou2021devil} from pre-trained depth estimation models. For example, Pseudo-LiDAR \cite{wang2019pseudo} generates the pseudo LiDAR information from the monocular image and depth information. CaDDN~\cite{reading2021categorical} projects LiDAR point clouds into
the image to create depth maps, then a categorical depth distribution is learned. Besides the depth and LiDAR data, there are also methods that try to utilize the CAD models~\cite{liu2021autoshape,manhardt2019roi,chabot2017deep,murthy2017reconstructing} to simplify the recognition and pose estimation of objects.  

The second kind of method aim to detect 3D objects without any extra data~\cite{mousavian20173d,liu2019deep,qin2019monogrnet,li2019gs3d,brazil2019m3d}. For example, M3D-RPN~\cite{brazil2019m3d} uses depth-aware convolutional layers to detect 3D objects. MonoPair~\cite{chen2020monopair} proposes to use the pairwise spatial relationships to achieve better results. SMOKE~\cite{liu2020smoke} proposes a CenterNet-style~\cite{zhou2019objects} 3D detector via keypoints estimation. Then, MonoFlex~\cite{MonoFlex} proposes a flexible center definition that unifies truncated objects and regular objects and an uncertainty-based depth ensemble method. To better find the bottleneck of purely monocular detectors, MonoDLE~\cite{ma2021delving} examines the effects of each component in mainstream methods. GrooMeD-NMS~\cite{kumar2021groomed} introduces differential non-maximal suppression into the monocular 3D object detection. MonoRUn~\cite{chen2021monorun} proposes to use the dense correspondence between 2D and 3D space to learn the reconstruction and reproject processes. To better utilize the geometry relationship in 3D and 2D space accompanying uncertainty, GUPNet~\cite{lu2021geometry} proposes a geometry uncertainty projection method to reduce the error in depth estimation.

\paragraph{Ground Plane Knowledge in the Monocular 3D Object Detection}
There have been several attempts in using the ground plane knowledge in monocular 3D object detection. Mono3D~\cite{chen2016monocular} first tries to use the ground plane to generate 3D bounding box proposals. Besides, Ground-Aware~\cite{liu2021ground} introduces the ground plane in the geometric mapping and proposes a ground-aware convolution module to enhance the detection.

In the above works, the ground plane is defined based on a fixed rule, that the camera height on KITTI is 1.65 meters. In this way, all positions at a fixed height of -1.65 meters in the 3D space would construct the ground plane. Different from the strong hypothesis of the fixed ground plane at the height of -1.65 meters, in this work, we propose a learnable ground plane prior that is based on a more reasonable hypothesis: objects should lie on the ground. As long as the objects lie on the ground, the ground plane can be substituted with the bottom surface of object's 3D bounding box. In this way, the proposed ground plane prior with the objects' bottom surface is object-wise adaptive and precise.

\paragraph{Depth Estimation in the Monocular 3D Object Detection}
There are two kinds of mainstream depth estimation methods in the purely monocular 3D object detection, which are direct regression~\cite{zhou2019objects} and geometry depth~\cite{cai2020monocular} derived from the pinhole imaging model. Further, MonoFlex~\cite{MonoFlex} extends the geometry depth to the diagonal keypoint depth by averaging the geometry depth of diagonal paired keypoints. Meanwhile, many works~\cite{MonoFlex,lu2021geometry,Shi_2021_ICCV,chen2021monorun} adopt the uncertainty along with the depth estimation to get better results.

In this work, with the help of the proposed ground plane prior, we propose a novel depth estimation method different from the above methods. It is a two-stage depth inference method that could get a more precise depth estimation compared with the geometry depth. Moreover, the proposed depth estimation can also adopt uncertainty and be extended with diagonal paired keypoints.
\section{Method}
\label{sec_method}
In this section, we elaborate on the details of our method. First, we give a brief problem definition of monocular 3D object detection. Second, we show how to utilize the ground plane prior and generate dense grounded depths. Third, the depth-align training strategy and two-stage depth inference method based on the ground plane prior are discussed. 

\subsection{Problem Definition}
\label{sub_sec_problem_definition}
The monocular 3D object detection task is to detect 3D objects from monocular RGB images. Besides the RGB image, calibration information (camera parameter matrix) can also be adopted. Specifically, for each object, the 3D location $(x,y,z)$, size $(h,w,l)$, and orientation $\theta$ are required. On the KITTI~\cite{Geiger2012CVPR} dataset, pitch and roll angles are considered as zero, so only orientation $\theta$ is considered.

Corresponding to the above targets, mainstream methods ~\cite{liu2020smoke, ma2021delving, MonoFlex} divide the whole task into four subtasks, which are 2D location, depth, size, and orientation estimation tasks. As discussed in \cref{sec_intro} and pointed out in \cite{ma2021delving}, depth estimation is the key bottleneck for monocular 3D object detection. In this way, we focus on precise depth estimation in this work.

\subsection{Ground Plane Prior}

\begin{figure}
    \centering
    \includegraphics[width=1.0\linewidth]{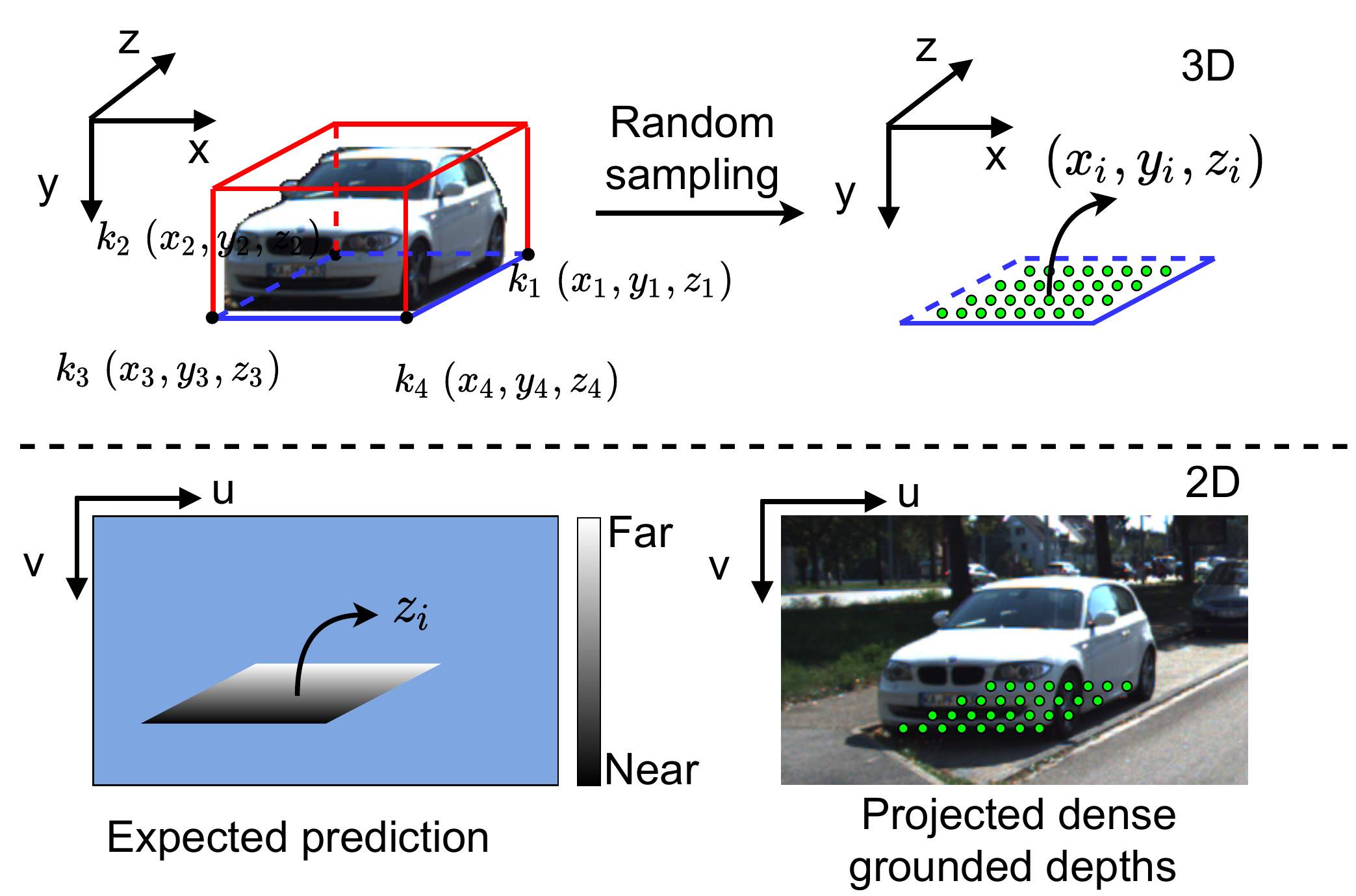}
    \caption{Illustration of the dense grounded depths. We first get the bottom of object's 3D bounding box, \ie, the ground plane. A random sampling is conducted on the plane, and the sampled results are projected to the image space to get dense grounded depths.}
    \label{fig_dense_depth}
    \vspace{-5pt}
\end{figure}

To define the ground plane, we start from a reasonable hypothesis that objects lie on the ground. This hypothesis holds for all common objects like cars, pedestrians, and cyclists. With this hypothesis, the ground plane can be substituted with the bottom surface of objects' 3D bounding box. For each object, we first get the bottom keypoints $\boldsymbol{k_1}, \boldsymbol{k_2}, \boldsymbol{k_3}$, and $\boldsymbol{k_4}$ in the 3D space. Then we conduct random sampling and interpolation on the 3D bottom surface composed of the bottom keypoints. Denote $R\in \mathbb{R}^{N\times 2}$ as a random matrix with each element $R_{ij} \in [0,1]$. $N$ is the number of sampling points. The sampled dense points can be written as:
\begin{equation}
    P_{3d} = R \begin{bmatrix}
        \boldsymbol{k_2} - \boldsymbol{k_1} \\
        \boldsymbol{k_4} - \boldsymbol{k_1}
    \end{bmatrix} + \boldsymbol{k_1}, 
\end{equation}
in which $P_{3d} \in \mathbb{R}^{N \times 3}$ contains all sampled points. Suppose $K_{3\times 3}$ is the camera parameter matrix. After the sampling, these points are projected back to image space:
\begin{equation}
    P_{2d}^T = K_{3\times 3} \ P_{3d}^T,
    \label{eq_3d_2d}
\end{equation}
in which $P_{2d} \in \mathbb{R}^{N \times 3}$. The $i$-th row of $P_{2d}$ is $[u_i \cdot z_i, v_i \cdot z_i, z_i]$, in which $u_i$ and $v_i$ are the projected coordinates in the image space, and $z_i$ is the corresponding depth at this point. The illustration is shown in \cref{fig_dense_depth}.

With the above formulation, we can get a dense and grounded depth representation, which addresses the problem of sparse depth supervision in mainstream methods. For the ill-posed 2D to 3D mapping problem, the dense grounded depth provides the ground plane constraint and makes it a well-posed problem, and the unique solution to the mapping is the intersection of camera ray and ground plane. Moreover, the irrelevant correlation problem is naturally avoided since our formulation has no relation to the object's height. More importantly, the introduced ground plane prior and dense grounded depth are derived from the 3D bounding box annotation, which can be easily accessed and requires no other data sources like LiDAR and depth.


\begin{figure*}
    \centering
    \includegraphics[width=0.8\linewidth]{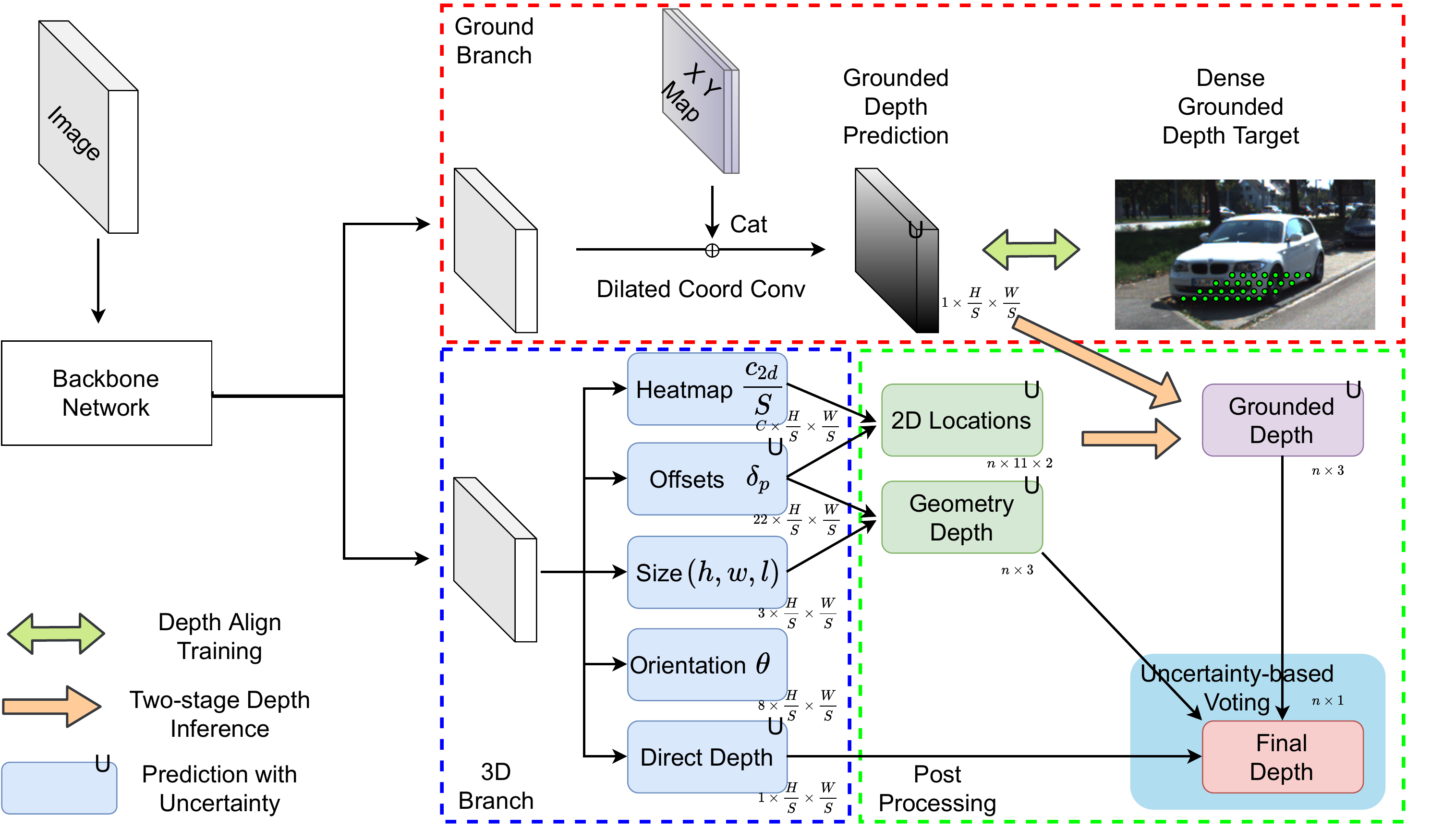}
    \caption{Framework design of MonoGround. There are three main parts, which are ground branch, 3D branch and post processing. In the post processing, $n$ is the number of objects.}
    \label{fig_framework}
    \vspace{-10pt}
\end{figure*}

\subsection{Detecting Objects with Dense Grounded Depth}
\label{sec_de_ob_with_dense_gd}
With the dense grounded depth, a natural question is how to use the ground plane prior to help the detecting of 3D objects. To answer this question, we start from three aspects in this section, which are network design, depth-align training, and two-stage depth inference.
\paragraph{Network Design} The framework of our method is shown in \cref{fig_framework}. We use DLA-34~\cite{DLA} as our backbone network. Then two branches are adopted, which are the ground branch and the 3D branch. The ground branch is to predict the grounded dense depth. The 3D branch is based on CenterNet~\cite{zhou2019objects} and is to predict attributes like location, size, orientation, \etc.

\begin{figure}
    \vspace{-10pt}
    \centering
    \includegraphics[width=1.0\linewidth]{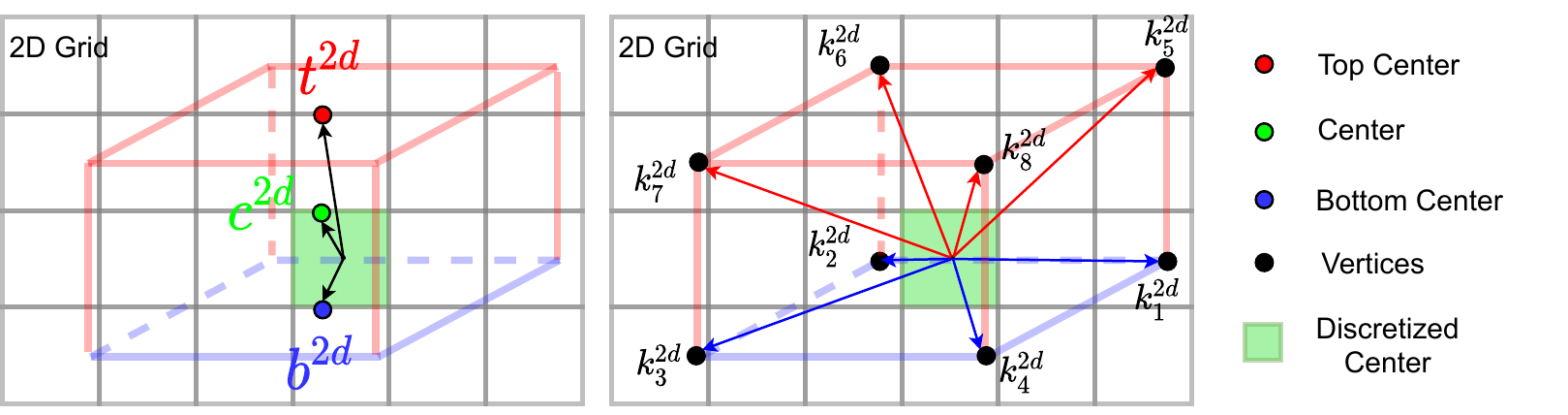}
    \caption{Illustration of the regression targets of the offset map.}
    \label{fig_11pts}
    \vspace{-10pt}
\end{figure}

In the 3D branch, we use MonoFlex~\cite{MonoFlex} as our baseline and simultaneously predict five kinds of outputs. The first one is the heatmap, which is used to represent and distinguish each object using its center. We follows the same definition of object center $\boldsymbol{c^{2d}}$ as MonoFlex~\cite{MonoFlex}. Heatmap can only predict rough locations due to quantization error. To get precise locations, we use offset maps to regress the fine-grained locations. Besides regressing the object center, we also use the offset map to regress other ten points, which are eight projected 3D bounding box vertices $\{\boldsymbol{k_1^{2d}}, \boldsymbol{k_2^{2d}}, \cdots, \boldsymbol{k_8^{2d}}\}$ and two projected bottom $\boldsymbol{b^{2d}}$ and top $\boldsymbol{t^{2d}}$ centers. The regression target of the offset map is shown in \cref{fig_11pts}, and we have:
\begin{equation}
    \label{eq_11pts}
    \begin{aligned}
        & \boldsymbol{\delta_{p}}=\dfrac{\boldsymbol{p}}{S} -  \lfloor \dfrac{\boldsymbol{c^{2d}}}{S} \rfloor, \\
        s.t. \ \ \  \boldsymbol{p} \in & \{\boldsymbol{c^{2d}},\boldsymbol{k_1^{2d}}, \cdots, \boldsymbol{k_8^{2d}}, \boldsymbol{b^{2d}}, \boldsymbol{t^{2d}}\},
    \end{aligned}
\end{equation}
in which $S$ is the output stride of the heatmap, and $\lfloor \cdot \rfloor$ is the floor function. Suppose $\boldsymbol{\delta^{*}_{p}}$ is the ground truth, the loss function of the offset map can be written as:
\begin{equation}
    L_{offset} = \sum_{\scriptsize \boldsymbol{p} \in \{\boldsymbol{c^{2d}},\boldsymbol{k_1^{2d}}, \cdots, \boldsymbol{k_8^{2d}}, \boldsymbol{b^{2d}}, \boldsymbol{t^{2d}}\} } \left | \boldsymbol{\delta_{p}} - \boldsymbol{\delta^{*}_{p}} \right |.
\end{equation}
Moreover, we also predict the size, orientation, and direct depth maps in the 3D branch as the baseline MonoFlex.

In the ground branch, we use dilated coord convolutions to get the final prediction of grounded depth. The reason why we use dilated convolution is that the predicted grounded depth is mostly occluded by the object, as shown in \cref{fig_dense_depth}. So we use dilated convolution to expand the receptive field of the network, and thus the surrounding ground plane can be seen by the network. In this way, the grounded depth can be better estimated. Moreover, coord convolutions~\cite{liu2018coordconv} are also adopted for the following two reasons. First, the estimation of grounded depth is related to both the visual and positional features. Plain convolution can only provide visual features, while coord convolution can integrate both visual and positional features.
Second, from \cref{eq_3d_2d} we can see that the 3D coordinates $(x,y,z)$ ($z$ is the depth) is a linear transformation of 2D coordinates. The coord convolutions can explicitly concatenate the 2D coordinates in the convolution and make it an external hint to realize a better estimation of depth. 

\begin{figure}
    \centering
    \begin{subfigure}{0.9\linewidth}
    \centering
    \includegraphics[width=0.8\linewidth]{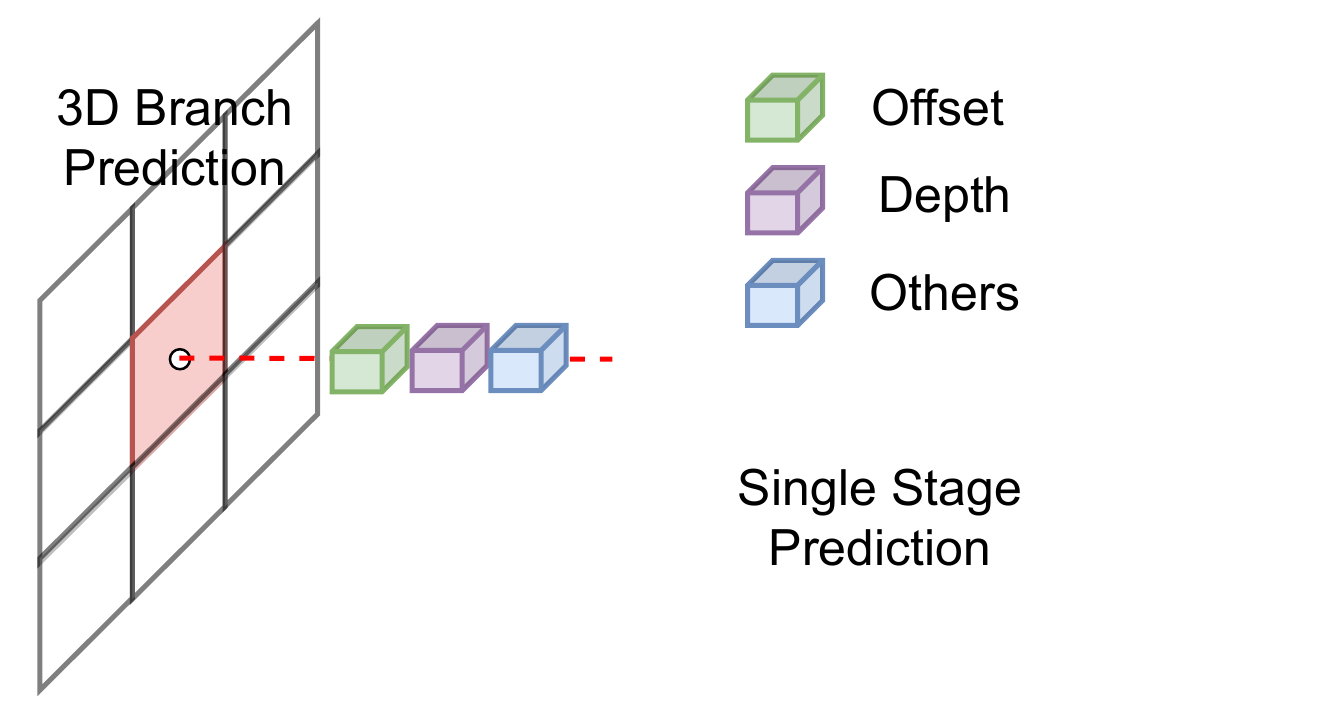}
    \caption{CenterNet-style depth inference.}
    \label{fig_twostage1}
    \end{subfigure}
    \begin{subfigure}{0.9\linewidth}
    \centering
    \includegraphics[width=0.8\linewidth]{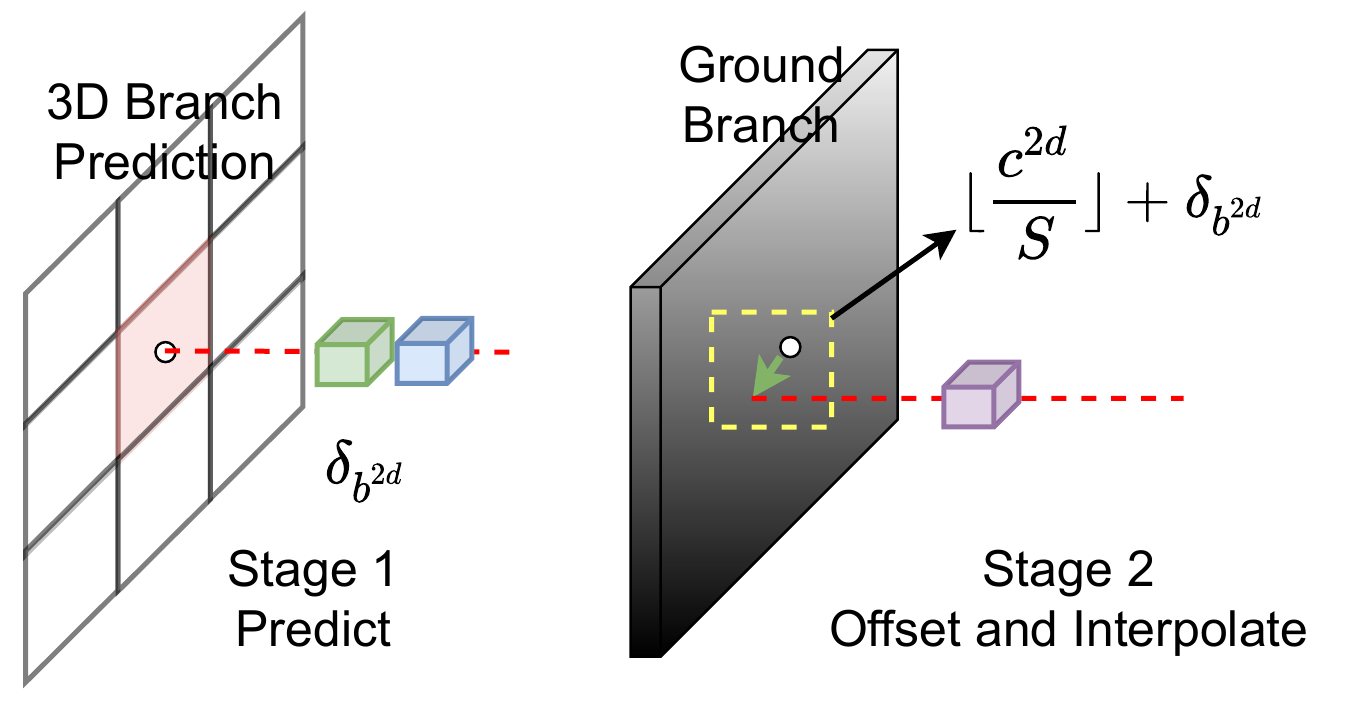}
    \caption{The proposed two-stage depth inference.}
    \label{fig_twostage2}
    \end{subfigure}
    \caption{Illustration and comparison of different depth inference methods.}
    \label{fig_twostage}
    \vspace{-5pt}
\end{figure}

In the post processing, we first get the 2D locations of the 11 keypoints as \cref{eq_11pts}, then we get the geometry depth~\cite{cai2020monocular}:
\begin{equation}
    z = \dfrac{f \cdot h}{h_{2d}},
\end{equation}
where $f$ is the focal length, $h$ is the predicted object height, and $h_{2d}$ is the pixel height obtained from 2D locations $\boldsymbol{b^{2d}}$ and $\boldsymbol{t^{2d}}$. We also extend the geometry depth by averaging the geometry depth of diagonal keypoints as~\cite{MonoFlex}. At last the final depth is obtained by an uncertainty-based voting from seven depths $z_{pred}$ (1 direct depth, 3 geometry depths, 3 grounded depths that will be discussed in \cref{fig_twostage} and the two-stage depth inference paragraph):
\begin{equation}
    z_{final} = \left( \sum_{i=1}^7 \dfrac{z_{pred}^i}{\sigma_i} \right) / \left(\sum_{i=1}^7 \dfrac{1}{\sigma_i} \right),
\end{equation}
where $\sigma_i$ is the predicted uncertainty along with the depth.

\begin{figure}
    \centering
    \includegraphics[width=0.6\linewidth]{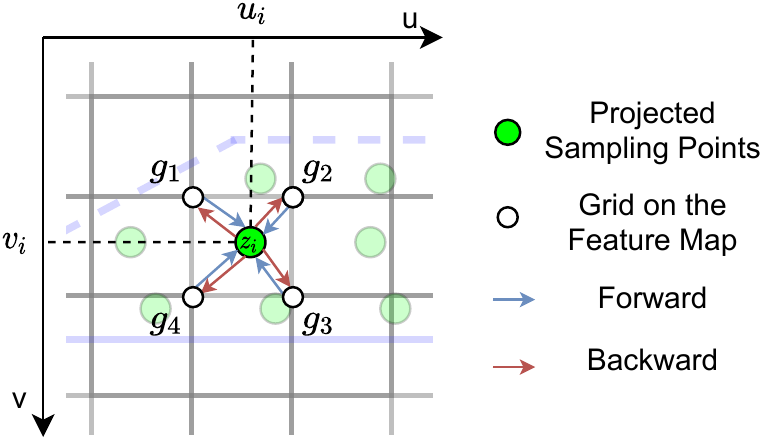}
    \caption{Demonstration of the depth-align training.}
    \label{fig_depth_align}
    \vspace{-8pt}
\end{figure}

\paragraph{Depth-Align Training}
After obtaining the prediction map of grounded depth, how to train this map is a major problem. There exists an issue of misalignment between the prediction map and projected dense grounded depths. The projected dense grounded points are scattered across the prediction map with arbitrary locations, while the prediction map is composed of uniform grids. In other words, the prediction map can only work with integer locations, while the projected dense grounded points are distributed in non-integer locations. To solve this problem, we propose a depth-align training method, as shown in \cref{fig_depth_align}.

The forward calculation of depth-align training is the same as the bilinear interpolation. Suppose the projected grounded point is $(u_i, v_i)$ with a depth of $z_i$, and $\{g_1, g_2, g_3, g_4\}$ are the four clockwise surrounding grid points from the left upper corner. Then the forward calculation can be written as:
\begin{equation}
    \begin{aligned}
        & \lambda_1 = u_i - \lfloor u_i \rfloor,\  \lambda_2 = \lceil u_i \rceil - u_i, \\
        & \lambda_3 = v_i - \lfloor v_i \rfloor,\ \  \lambda_4 = \lceil v_i \rceil - v_i, \\
        pred_i = &\lambda_2 \lambda_4 g_1 + \lambda_1 \lambda_4 g_2 + \lambda_1 \lambda_3 g_3 +  \lambda_2 \lambda_3 g_4, 
    \end{aligned}
    \label{eq_interp}
\end{equation}
in which $\lfloor \cdot \rfloor$ and $\lceil \cdot \rceil$ are the floor and ceiling functions, and $pred_i$ is the prediction of depth. Then the loss can be:
\begin{equation}
    L_{da} = \sum_{i=1}^N \left | z_i - pred_i \right |.
    \label{eq_da_loss}
\end{equation}
The variables $z_i$, $u_i$, and $v_i$ are obtained from the $P_{2d}$ in \cref{eq_3d_2d}.
In this way, we can directly optimize the dense depths with non-integer locations, and the backward calculation is to pass the gradient to the surrounding grids according to the weights of the bilinear interpolation. 

\paragraph{Two-stage Depth Inference}
In CenterNet-style~\cite{zhou2019objects} inference, objects are represented by coarse grid points on the feature map, and all other attributes are predicted from the features at the points at once. In this work, we call this one-stage inference. However, the coarse grid points can only reflect the rough location of objects, and this is why we need to further regress the difference between ground truth locations and rough grid locations. 

With the help of the densely trained grounded depth map, we can achieve more precise depth estimation compared with the original CenterNet-style depth inference by using the regressed fine-grained locations instead of rough grid locations. Suppose $\lfloor \boldsymbol{c^{2d}} / S \rfloor$ is the center predicted from the heatmap. We first apply the regressed offset to the center to obtain the fine location $\boldsymbol{b^{2d}}/ S = \lfloor \boldsymbol{c^{2d}} / S \rfloor + \boldsymbol{\delta_{b^{2d}}}$, then we use the interpolation as \cref{eq_interp} on the prediction of ground branch to get the precise depth. Since $\boldsymbol{b^{2d}}$ is the bottom center of the object, the obtained grounded depth of $\boldsymbol{b^{2d}}$ is exactly the same as the depth of $\boldsymbol{c^{2d}}$, \ie, the depth of object center. The illustration and comparison of different depth inference methods are shown in \cref{fig_twostage}.

Similarly, we can obtain precise grounded depth for the keypoints $\{ \boldsymbol{k_1^{2d}}$, $\boldsymbol{k_2^{2d}}, \boldsymbol{k_3^{2d}}$, $\boldsymbol{k_4^{2d}} \}$. According to the 3D geometry, the center depth equals to the average depth of the diagonal keypoints. In this way, we can get another two depth estimations from keypoints $\boldsymbol{k_1^{2d}}, \boldsymbol{k_3^{2d}}$ and $\boldsymbol{k_2^{2d}}, \boldsymbol{k_4^{2d}}$ by averaging the obtained diagonal results. Since we utilize the regression results before conducting depth estimation, we call this two-stage depth inference.

It is worth mentioning that this kind of precise two-stage depth inference can only be carried out with the help of the proposed ground plane prior because this process needs the ability to obtain depths from any given non-integer location, \ie, interpolating across densely trained \textbf{grounded depth} maps. Previous works are mainly designed to cope with the sparse and non-grounded depth supervision, which is hard to adopt a similar two-stage depth inference.

\section{Experiments}
\label{sec_exp}
\subsection{Implementation Details}
\paragraph{Dataset\protect\footnote{We also add the experiments on NuScenes~\cite{Caesar_2020_CVPR} in our open-source code repository.}}
In this work, we use KITTI~\cite{Geiger2012CVPR} vision benchmark to evaluate and test our method. KITTI vision benchmark is a multimodal and multi-task dataset. We use the data from the 3D object detection track, which contains 7481 training images and 7518 testing images. Besides the RGB images and camera calibration matrices, no other data is utilized in our work. We also follow the split from~\cite{chen20153d} which divide the whole 7481 training images into a \textit{train} set (3712) and a \textit{val} set (3769).
\paragraph{Metrics}
We use the 3D average precision with 40 recall positions~\cite{simonelli2019disentangling} AP$_{3D|R40}$ under three different difficulties (easy, moderate, hard) as our main evaluation metrics. Bird-eye view (BEV) detection is also adopted as our evaluation metrics. To better compare the performance of depth estimation, we also use the mean percentage error (MPE) as our metric:
\begin{equation}
    MPE = \sum_i \left | \dfrac{pred_i - gt_i}{gt_i} \right |.
\end{equation}

\paragraph{Training Details}
The size of the input image is padded to 384$\times$1280, and random horizontal flip augmentation is utilized. In the ground branch, we use two successive 3$\times$3 Conv-BN-ReLU layers and one output convolution layer. All convolutions in the ground branch are coord convolutions~\cite{liu2018coordconv}, and their dilations are set to 2. For each object, the number of sampling points of dense grounded depth is the same as the polygon area of the projected bottom quadrilateral. The biggest number of sampling points does not exceed 5500 on KITTI.

We use PyTorch~\cite{NEURIPS2019_9015} to implement our method and train it with RTX 2080Ti GPU. AdamW~\cite{loshchilov2018decoupled} optimizer is adopted with a learning rate of 3e-4 and a weight decay of 1e-5. The batch size is set to 8, and the number of training epochs is set to 100. The learning rate will be decayed by a factor of 0.1 at the 80th and the 90th epochs.

\subsection{Ablation Study}
\begin{figure}[h]
    \centering
    \includegraphics[width=0.9\linewidth]{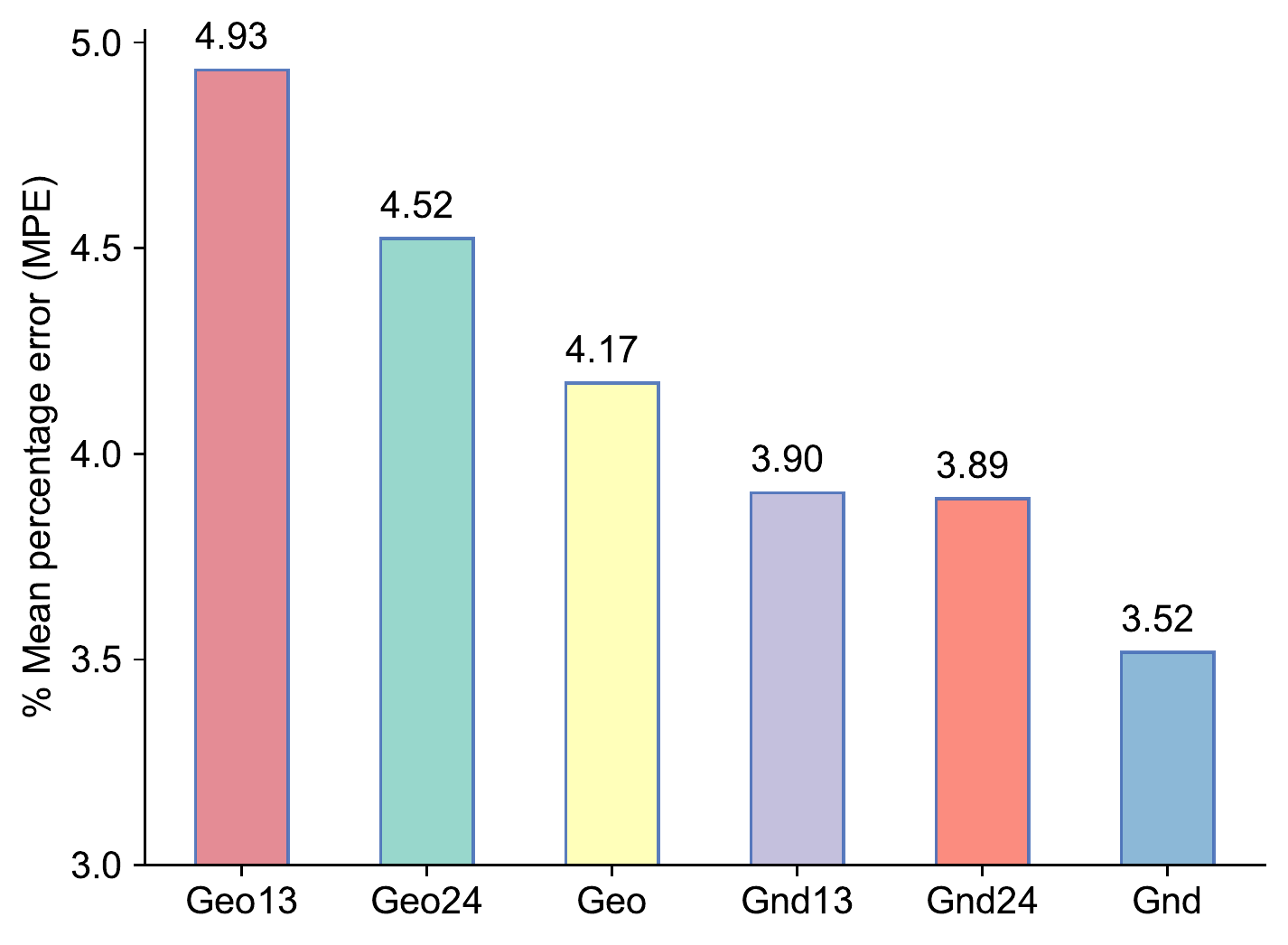}
    \vspace{-3pt}
    \caption{The mean percentage errors (MPEs) of the geometry depths and the grounded depths (lower is better). }
    \label{fig_different_depth}
    \vspace{-2pt}
\end{figure}

In this section, we discuss our method from two aspects, which are the effectiveness of the grounded depth, the ablation of each component in the ground branch.
\paragraph{Effectiveness of the Grounded Depth}
To examine the effectiveness of the grounded depth, we compare the MPE of our method and the widely used geometry depth estimation methods. The results are shown in \cref{fig_different_depth}. ``Geo" means using the geometry depth estimation, and ``Gnd" means using the grounded depth. The postfixes ``13" and ``24" means using the average depths from diagonal keypoints $\boldsymbol{k_1}, \boldsymbol{k_3}$ and $\boldsymbol{k_2}, \boldsymbol{k_4}$, respectively.

From \cref{fig_different_depth}, we can see that the proposed grounded depth outperforms the widely used geometry depth in all settings. This shows the effectiveness of the grounded depth and the idea of introducing the ground plane prior.

\vspace{-5pt}
\paragraph{Ablation Study on the Ground Branch}
To verify the effectiveness of the design of the ground branch, we show the ablation study in this section, as shown in \cref{tab_ablation}. 

\begin{table}[h]
    \centering
    \caption{Ablation study on the KITTI \textbf{\textit{val}} set. We use the average AP$_{3D|R40}$ with a IOU threshold of 0.7 is used as the evaluation metric. ``sp." means adding the grounded depth branch, but with the sparse supervision as the conventional depth training. ``de." means adding the dense grounded depth branch. ``co." means the coord convolution is adopted. ``di." means using the dilated convolution.}
    \label{tab_ablation}
    \resizebox{0.9\linewidth}{!}{%
    \begin{tabular}{lccc}
    \toprule
    Setting              & Easy  & Moderate & Hard \\
    \midrule
    Baseline             & 22.18  & 16.37     & 13.98 \\
    Baseline+sp.         & 19.23  & 14.14     & 11.84 \\
    Baseline+de.         & 20.54 & 15.54     & 13.34 \\
    Baseline+de.+co.     & 21.54  & 16.23     & 13.84 \\
    Baseline+de.+di.     & 22.65  & 16.67     & 14.37 \\
    Baseline+de.+co.+di. & 22.98 & 17.37     & 14.78 \\
    \bottomrule
    \end{tabular}}
\end{table}

We can see that the introduced grounded depth with dense training outperforms that with the sparse supervision. Moreover, the coord and dilated convolution also gains better performance as discussed in \cref{sec_de_ob_with_dense_gd}.

\begin{table*}
    \caption{Comparison on the KITTI \textbf{\textit{test}} set for the car class. The results are tested on the KITTI testing server.} 
    \label{tb_kitti_test}
    \centering
    \resizebox{0.9\linewidth}{!}{%
    \begin{tabular}{lcccccc@{\hskip 0.3in}ccc}
    \toprule
    \multirow{2}{*}{Method} & \multirow{2}{*}{Year}    & Extra        & Runtime & \multicolumn{3}{c}{AP$_{3D|R40}$}                     & \multicolumn{3}{c}{BEV} \\ 
    &&Data&(ms)& easy & moderate & hard & easy & moderate & hard \\
    \midrule
    Decoupled-3D~\cite{cai2020monocular}               & AAAI20  & Depth        & -       & 11.08  & 7.02  & 5.63  & 23.16  & 14.82  & 11.25 \\
    MonoPSR~\cite{ku2019monocular}                      & CVPR19 & LiDAR           & 200      & 10.76  & 7.25  & 5.85  & 18.33  & 12.58  & 9.91 \\
    AM3D~\cite{ma2019accurate}                       & ICCV19  & Depth        & ~400 & 16.50  & 10.74 & 9.52  & 25.03  & 17.32  & 14.91 \\
    PatchNet~\cite{ma2020rethinking}                   & ECCV20  & Depth        & ~400 & 15.68  & 11.12 & 10.17 & 22.97  & 16.86  & 14.97 \\
    DA-3Ddet~\cite{ye2020monocular}                   & ECCV20  & Depth        & - & 16.80  & 11.50 & 8.90 & -  & -  & - \\
    D4LCN~\cite{ding2020learning}                      & CVPR20  & Depth        & -       & 16.65  & 11.72 & 9.51  & 22.51  & 16.02  & 12.55 \\
    Kinem3D~\cite{brazil2020kinematic}                    & ECCV20  & Multi-frames & 120     & 19.07  & 12.72 & 9.17  & 26.69  & 17.52  & 13.10 \\
    PCT~\cite{wang2021progressive}                       & NeurIPS21  & Depth           & 45     & 21.00  & 13.37 & 11.31  & 29.65  & 19.03  & 15.92 \\
    CaDDN~\cite{reading2021categorical}                      & CVPR21  & LiDAR        & 63      & 19.17  & 13.41 & 11.46 & 27.94  & 18.91  & 17.19 \\
    DFR-Net~\cite{zou2021devil}                       & ICCV21  & Depth           & 180     & 19.40  & 13.63 & 10.35  & 28.17  & 19.17  & 14.84 \\
    AutoShape~\cite{liu2021autoshape}                  & ICCV21  & CAD Models   & 50      & 22.47  & 14.17 & 11.36 & 30.66  & 20.08  & 15.59 \\ \midrule
    M3D-RPN~\cite{brazil2019m3d}                      & ICCV19 & No           & 160      & 14.76  & 9.71  & 7.42  & 21.02  & 13.67  & 10.23 \\
    SMOKE~\cite{liu2020smoke}                      & CVPRW20 & No           & 30      & 14.03  & 9.76  & 7.84  & 20.83  & 14.49  & 12.75 \\
    MonoPair~\cite{chen2020monopair}                   & CVPR20  & No           & 57      & 13.04  & 9.99  & 8.65  & 19.28  & 14.83  & 12.89 \\
    MonoDLE~\cite{ma2021delving}                    & CVPR21  & No           & 40      & 17.23  & 12.26 & 10.29 & 24.79  & 18.89  & 16.00 \\
    MonoRUn~\cite{chen2021monorun}                    & CVPR21  & No           & 70      & 19.65  & 12.30 & 10.58 & 27.94  & 17.34  & 15.24 \\
    GrooMeD~\cite{kumar2021groomed}                & CVPR21  & No           & 120     & 18.10  & 12.32 & 9.65  & 26.19  & 18.27  & 14.05 \\
    MonoRCNN~\cite{Shi_2021_ICCV}                  & ICCV21  & No           & 70     & 18.36  & 12.65 & 10.03  & 25.48  & 18.11  & 14.10 \\
    DDMP-3D~\cite{wang2021depth}                    & CVPR21  & No           & 180     & 19.71  & 12.78 & 9.80  & 28.08  & 17.89  & 13.44 \\
    MonoEF~\cite{zhou2021monocular}                     & CVPR21  & No           & 30      & 21.29  & 13.87 & 11.71 & 29.03  & 19.70  & 17.26 \\
    MonoFlex~\cite{MonoFlex}                   & CVPR21  & No           & 30      & 19.94  & 13.89 & 12.07 & 28.23  & 19.75  & 16.89 \\ 
    GUPNet~\cite{lu2021geometry}                   & ICCV21  & No           & 30      & 20.11  & 14.20 & 11.77 & -  & -  & - \\ 
    MonoGround                 &         & No           & 30      & \textbf{21.37}  & \textbf{14.36} & \textbf{12.62} & \textbf{30.07}  & \textbf{20.47}  & \textbf{17.74} \\ \bottomrule
    \end{tabular}}
\end{table*}
\begin{table*}
\centering
\caption{Comparison on the KITTI \textbf{\textit{val}} set for the car class. * means we use the results from the official open-source code, which are slightly different from the reported results.}
\label{tb_kitti_val}
\resizebox{0.9\linewidth}{!}{%
\begin{tabular}{lc@{\hskip 0.07in}c@{\hskip 0.07in}c@{\hskip 0.2in}c@{\hskip 0.07in}c@{\hskip 0.07in}c@{\hskip 0.2in}c@{\hskip 0.07in}c@{\hskip 0.07in}c@{\hskip 0.2in}c@{\hskip 0.07in}c@{\hskip 0.07in}c}
\toprule
\multirow{2}{*}{Method}      & \multicolumn{3}{c@{\hskip 0.2in}}{AP$_{3D|R40}$@IOU=0.7} &\multicolumn{3}{c@{\hskip 0.2in}}{BEV@IOU=0.7} & \multicolumn{3}{c@{\hskip 0.2in}}{AP$_{3D|R40}$@IOU=0.5} & \multicolumn{3}{c@{\hskip 0.2in}}{BEV@IOU=0.5} \\ 
& easy & moderate & hard & easy & moderate & hard & easy & moderate & hard& easy & moderate & hard                  
\\ \midrule
CenterNet~\cite{zhou2019objects}   & 0.60    & 0.66    & 0.77    & 3.46   & 3.31   & 3.21   & 20.00   & 17.50   & 15.57   & 34.36  & 27.91  & 24.65  \\
MonoGRNet~\cite{qin2019monogrnet}   & 11.90   & 7.46    & 5.76    & 19.72  & 12.81  & 10.15  & 47.59   & 32.28   & 25.50   & 48.53  & 35.94  & 28.59  \\
MonoDIS~\cite{simonelli2019disentangling}     & 11.06   & 7.60    & 6.37    & 18.45  & 12.58  & 10.66  & -       & -       & -       & -      & -      & -      \\
M3D-RPN~\cite{brazil2019m3d}     & 14.53   & 11.07   & 8.65    & 20.85  & 15.62  & 11.88  & 48.53   & 35.94   & 28.59   & 53.35  & 39.60  & 31.76  \\
MoVi-3D~\cite{simonelli2020towards}     & 14.28   & 11.13   & 9.68    & 22.36  & 17.87  & 15.73  & -       & -       & -       & -      & -      & -      \\
MonoPari~\cite{chen2020monopair}    & 16.28   & 12.30   & 10.42   & 24.12  & 18.17  & 15.76  & 55.38   & 42.39   & 37.99   & 61.06  & 47.63  & 41.92  \\
MonoDLE~\cite{ma2021delving}     & 17.45   & 13.66   & 11.68   & 24.97  & 19.33  & 17.01  & 55.41   & 43.42   & 37.81   & 60.73  & 46.87  & 41.89  \\
GrooMeD~\cite{kumar2021groomed} & 19.67   & 14.32   & 11.27   & 27.38  & 19.75  & 15.92  & 55.62   & 41.07   & 32.89   & 61.83  & 44.98  & 36.29  \\
GUPNet~\cite{lu2021geometry}      & 22.76   & 16.46   & 13.72   & 31.07  & 22.94  & 19.75  & 57.62   & 42.33   & 37.59   & 61.78  & 47.06  & 40.88  \\
MonoFlex*~\cite{MonoFlex}    & 24.22   & 17.34   & 15.13   & 31.65  & 23.29  & 20.02  & 60.70   & 45.65   & 39.91   & 66.26  & 49.30  & 44.42  \\
MonoGround  & \textbf{25.24}   & \textbf{18.69}   & \textbf{15.58}   & \textbf{32.68}  & \textbf{24.79}  & \textbf{20.56}  & \textbf{62.60}   & \textbf{47.85}   & \textbf{41.97}   & \textbf{67.36}  & \textbf{51.83}  & \textbf{45.65} \\
\bottomrule
\end{tabular}}
\end{table*}

\subsection{Performance Evaluation}
\label{sec_perf_eval}
We report the evaluation results on the both KITTI \textit{test} and \textit{val} sets. The results for the car class are shown in \cref{tb_kitti_test,tb_kitti_val}. From \cref{tb_kitti_test} we can see that our method outperforms all other purely monocular methods while maintaining a fast speed in real-time. When compared with the methods using extra data sources, our method still gets the highest performance in the ``moderate" and ``hard" setttings. Similar results can also be seen in \cref{tb_kitti_val}. Moreover, we can see that our method has a relatively large performance gap compared with other methods. When the intersection over union (IOU) is set to 0.5, the performance gap becomes larger($\sim$2\%).

\begin{figure*}
    \centering
    \includegraphics[width=1.0\linewidth]{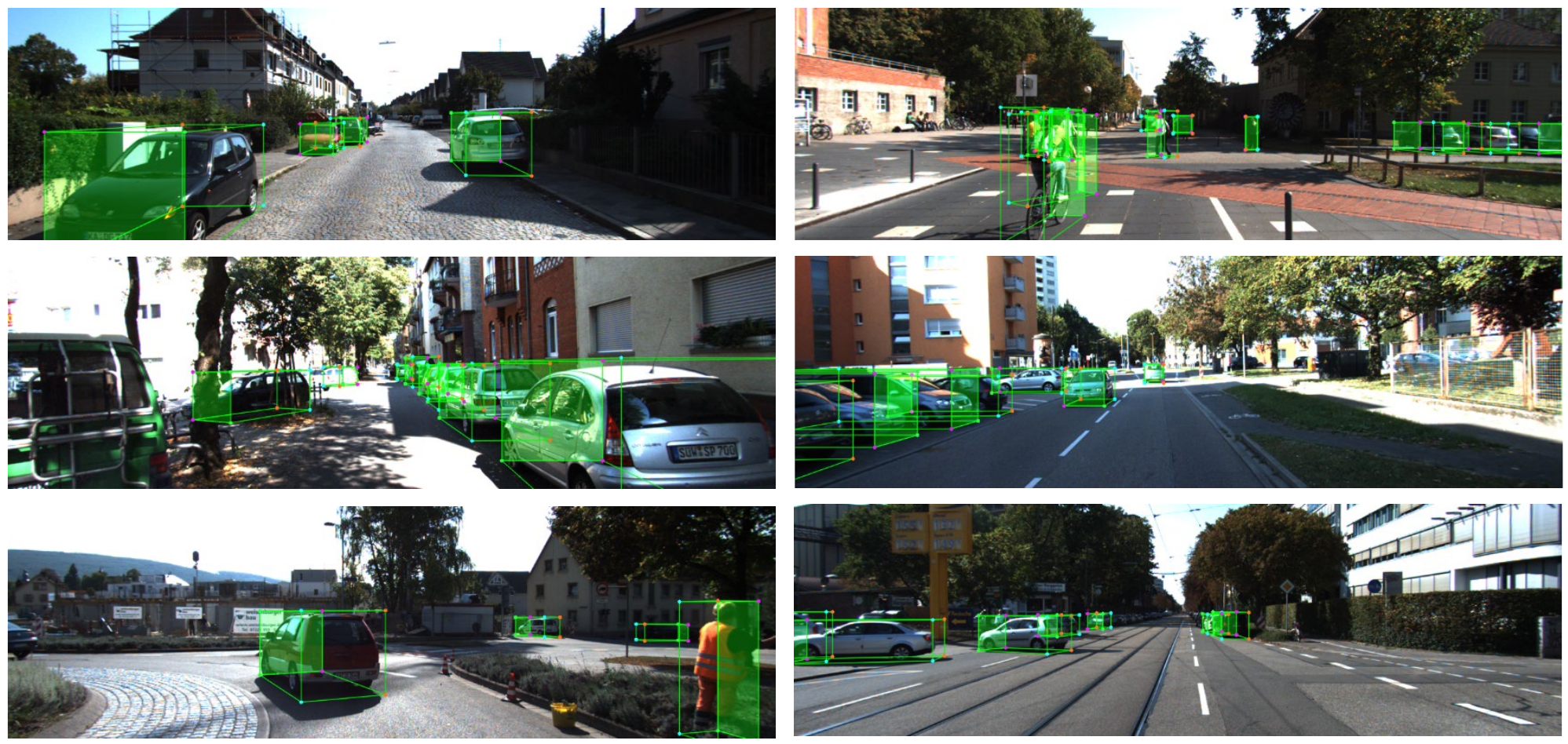}
    \caption{Visualization of the detection results on the KITTI \textbf{\textit{test}} set.}
    \label{fig_vis}
    \vspace{-10pt}
\end{figure*}

We also show the results for the pedestrian and cyclist classes on the KITTI \textit{test} set. As shown in \cref{tb_ped_cyc_test}, our method still performs well on these two classes. Our method obtains the best results for the cyclist class and the second-best results for the pedestrian class. A potential reason for the relatively low performance for the pedestrian class is that the pedestrians' sizes (width and length) are relatively small compared with cars and cyclists. In this way, the bottom rectangle of the 3D bounding box is small and the projected area on the ground is small. It would be hard to conduct random sampling on this small area as shown in \cref{fig_dense_depth}. So the performance increases for the pedestrian class is not as large as other classes. However, our method still outperforms other methods (except GUPNet) by a relatively big margin for the pedestrian class ($\sim$2\%, $\sim$1\%, and $\sim$1.5\% for the ``easy", ``moderate", and ``hard" settings).

\begin{table}[t]
\centering
\caption{The performance of the AP$_{3D|R40}$ on the KITTI \textbf{\textit{test}} set for the pedestrian and cyclist classes. ``mod." means moderate. }
\label{tb_ped_cyc_test}
\resizebox{0.95\linewidth}{!}{%
\begin{tabular}{lc@{\hskip 0.07in}c@{\hskip 0.07in}cc@{\hskip 0.07in}c@{\hskip 0.07in}c}
\toprule
\multirow{2}{*}{Method}   & \multicolumn{3}{c}{Pedestrian}  & \multicolumn{3}{c}{Cyclist}  \\
& easy & mod. & hard & easy & mod. & hard \\
\midrule
M3D-RPN~\cite{brazil2019m3d}    & 4.92  & 3.48 & 2.94 & 0.94 & 0.65 & 0.47 \\
DDMP-3D~\cite{wang2021depth}    & 4.93  & 3.55 & 3.01 & 4.18 & 2.50 & 2.32 \\
Movi3D~\cite{simonelli2020towards}     & 8.99  & 5.44 & 4.57 & 1.08 & 0.63 & 0.70 \\
MonoPair~\cite{chen2020monopair}   & 10.02 & 6.68 & 5.53 & 3.79 & 2.12 & 1.83 \\
MonoFlex~\cite{MonoFlex}   & 9.43  & 6.31 & 5.26 & 4.17 & 2.35 & 2.04 \\
MonoDLE~\cite{ma2021delving}    & 9.64  & 6.55 & 5.44 & 4.59 & 2.66 & 2.45 \\
MonoRUn~\cite{chen2021monorun}    & 10.88 & 6.78 & 5.83 & 1.01 & 0.61 & 0.48 \\
GUPNet~\cite{lu2021geometry}     & \textbf{14.7}2 & \textbf{9.53} & \textbf{7.87} & 4.18 & 2.65 & 2.09 \\
MonoGround & 12.37 & 7.89 & 7.13 & \textbf{4.62} & \textbf{2.68} & \textbf{2.53} \\
\bottomrule
\end{tabular}}
\end{table}

\subsection{Visualization}

In this section, we show the visualization results of the proposed method on the KITTI \textit{test} set, as shown in \cref{fig_vis}. We can see that our method gives good visualization results in detecting various 3D objects.

\section{Conclusion and Limitation}
\label{sec_conclusion}
In this work, we have proposed a monocular 3D object detection method termed as MonoGround, which introduces the ground plane prior in monocular 3D object detection. The introduced dense grounded depth with the ground plane prior requires no extra data sources like LiDAR, stereo images, and depth images. Moreover, we also have proposed a depth-align training strategy and a two-stage precise depth inference method to cope with the introduced dense grounded depth in the network. The proposed depth inference method can be easily extended with uncertainty and has outperformed the conventional geometry depth inference method. With these components, MonoGround has achieved state-of-the-art performance on KITTI while maintaining a very fast real-time speed.

However, there is a limitation in this work. As discussed in \cref{sec_perf_eval}, the introduced ground plane prior needs a random dense sampling process, which is less friendly for objects with a small bottom surface like pedestrians, since the sampled points and depths could be rare. Although we still get the second-best result for the pedestrian class, this could be a direction for the future work.

\vspace{-5pt}
\section*{Acknowledgements}
\vspace{-5pt}

This work is supported in part by National Key Research and Development Program of China under Grant 2020AAA0107400, Zhejiang Provincial Natural Science Foundation of China under Grant LR19F020004, National Natural Science Foundation of China under Grant U20A20222, and Zhejiang University K.P.Chao's High Technology Development Foundation.

{\small
\bibliographystyle{ieee_fullname}
\bibliography{egbib}
}

\end{document}